\documentclass{article}

\usepackage{microtype}
\usepackage{graphicx}
\usepackage{subfigure}
\usepackage{booktabs} 
\usepackage{amsmath,amssymb}
\usepackage{amsfonts, amsthm}
\usepackage{soul,color,xcolor}
\usepackage{multirow,multicol}
\usepackage{makecell}
\usepackage{float}

\usepackage[utf8x]{inputenc}
\usepackage{array,tabularx}
\usepackage{bm}
\usepackage{float}
\usepackage{placeins}
\usepackage{multibib}
\usepackage{wasysym}
\usepackage{xcolor}

\newcites{app}{References}

\usepackage{hyperref}

\usepackage[accepted]{icml2019}

\icmltitlerunning{Towards a More Complete Theory of Function Preserving Transforms}

\begin{document}

\twocolumn[
\icmltitle{Towards a More Complete Theory of Function Preserving Transforms}



\icmlsetsymbol{equal}{*}

\begin{icmlauthorlist}
\icmlauthor{Michael Painter}{}
\end{icmlauthorlist}


\icmlcorrespondingauthor{Michael Painter}{mpainter.thts@gmail.com}

\icmlkeywords{Machine Learning, ICML}

\vskip 0.3in
]



\newcommand{\li}[3]{{ 
    {#1}^{#2}_{#3}
}}
\newcommand{\la}[2]{{#1}^{#2}}

\newcommand{\ef}{\mathcal{F}}

\newcommand{\etal}{\textit{et al.}}

\newcommand{\R}{\texttt{R2R}}
\newcommand{\Rwider}{\texttt{R2WiderR}}
\newcommand{\Rdeeper}{\texttt{R2DeeperR}}

\newcommand{\N}{\texttt{Net2Net}}
\newcommand{\Nwider}{\texttt{Net2WiderNet}}
\newcommand{\Ndeeper}{\texttt{Net2DeeperNet}}

\newcommand{\NM}{\texttt{NetMorph}}

\newcommand{\zeroatzero}{Zero@Zero}

\DeclareRobustCommand{\add}[1]{{\sethlcolor{pink}\hl{#1}}}
\DeclareRobustCommand{\rm}[1]{{\sethlcolor{green}\hl{#1}}}
\DeclareRobustCommand{\unclear}[1]{{\sethlcolor{red}\hl{#1}}}

\newcommand{\netmorphauth}{Wei \textit{et al.}}

\printAffiliationsAndNotice{}  

\section*{Foreword} 

I completed the majority of this work for a project for the CS231n: Deep Learning for Computer Vision course at Stanford. The original premise of the project was to use the function-preserving transforms described below to perform an efficient architecture search. At the start of my PhD in 2018 I spent some time writing this work up for a conference submission but subsequently didn't find time to work on this further.

I submitted this work to ICML2019 and received scores of \texttt{\small\{Accept, Weak Accept, Weak Reject, Reject\}}, where the main strengths highlighted were the mathematical derivations and the main concerns had to do with the performance of the networks used in the evaluation. Appendix \ref{app:A} which links to code also briefly discusses the concerns from reviewers.

Additionally, I also found some plots from the original CS231n project report which are more precise (but not better in terms of performance) and possibly better demonstrate the potential computational speedup from using transformations during training. I've added them in Appendix \ref{app:B}.



Given recent developments using extremely large neural networks with billions and trillions of parameters, I'm guessing that there may be some interest in trying to train them more efficiently, so I'm publishing this work to arxiv on the odd chance that someone may find this work useful. The idea of using symmetry in the initialisation of new weights in a function-preserving transform should easily generalize to more recent attention and transformer architectures.

The updates I've made to this paper for this arxiv version relative to the version submitted to ICML are: 
(1) changing the paper format to be non-anonymous; 
(2) adding this foreword (and shifting the abstract, introduction and original paper
 to start on a clean page);
(3) adding a link to the code in Appendix \ref{app:A};
(4) added Appendix \ref{app:B}, with plots from the CS231n project report;
(5) a small comment about symmetry breaking in Appendix \ref{app:full_derivation}.
The rest is as is unfortunately.

\ 

\ 

\smiley{}

\textcolor{white}{
    I completed the majority of this work for a project for the CS231n: Deep Learning for Computer Vision course at Stanford. The original premise of the project was to use the function-preserving transforms described below to perform an efficient architecture search. At the start of my PhD in 2018 I spent some time writing this work up for a conference submission but subsequently didn't find time to work on this further.   
}
    
\textcolor{white}{
    I submitted this work to ICML2019 and received scores of \texttt{\small\{Accept, Weak Accept, Weak Reject, Reject\}}, where the main strengths highlighted were the mathematical derivations and the main concerns had to do with the performance of the networks used in the evaluation. Appendix A
    which links to code also briefly discusses the concerns from reviewers.   
}
    
\textcolor{white}{
    Additionally, I also found some plots from the original CS231n project report which are more precise (but not better in terms of performance) and possibly better demonstrate the potential computational speedup from using transformations during training. I've added them in Appendix B.
}
    
\textcolor{white}{
    Given recent developments using extremely large neural networks with billions and trillions of parameters, I'm guessing that there may be some interest in trying to train them more efficiently, so I'm publishing this work to arxiv on the odd chance that someone may find this work useful. The idea of using symmetry in the initialisation of new weights in a function-preserving transform should easily generalize to more recent attention and transformer architectures.   
}
    
\textcolor{white}{
    The updates I've made to this paper for this arxiv version relative to the version submitted to ICML are: 
    (1) changing the paper format to be non-anonymous; 
    (2) adding this foreword (and shifting the abstract, introduction and original paper
     to start on a clean page);
    (3) adding a link to the code in Appendix A;
    (4) added Appendix B, with plots from the CS231n project report;
    (5) a small comment about symmetry breaking in Appendix C.
    The rest is as is unfortunately. 
    (Also if you're reading this then I hope this joke on the transformation in the paper isn't lost on you!)
}

\clearpage
\begin{abstract}

In this paper, we develop novel techniques that can be used to alter the architecture of a neural network, while maintaining the function it represents. Such operations are known as function preserving transforms and have proven useful in transferring knowledge between networks to evaluate architectures quickly, thus having applications in efficient architectures searches. Our methods allow the integration of residual connections into function preserving transforms, so we call them \R. We provide a derivation for \R\ and show that it yields competitive performance with other function preserving transforms, thereby decreasing the restrictions on deep learning architectures that can be extended through function preserving transforms. We perform a comparative analysis with other function preserving transforms such as \N\ and Network Morphisms, where we shed light on their differences and individual use cases. Finally, we show the effectiveness of \R\ to train models quickly, as well as its ability to learn a more diverse set of filters on image classification tasks compared to \N\ and Network Morphisms.

\end{abstract}

\section{Introduction}

\textit{Function preserving transforms} (FPTs), also known as \textit{network morphisms}, provide a method for transferring the performance of a \textit{teacher network}, $f$, to a \textit{student network}, $g$, with a different (typically larger) architecture. This is assured by computing the initial parameters $\theta_s$ for the student network from the parameters of the teacher $\theta_t$, such that they are \textit{function preserving}: $\forall x.\ f(x;\theta_t) = g(x;\theta_s)$.

FPTs allow us to dynamically increase the capacity of a network, during training, without degrading performance. This enables us to leverage an already trained teacher network and perform a fast evaluation of new architectures, without incurring the overhead of training them from scratch. In particular, FPTs have applications in efficient architecture searches \cite{n2n_architecture_search,nm_architecture_search}. Many architectures can be evaluated quickly by repeatedly applying FPTs, thereby amortizing the high cost of training from a random initialization over many architecture evaluations.

Existing methods for performing FPTs include \N\ \cite{net2net} and Network Morphing \cite{netmorph}. \N\   provides techniques for \textit{widening} hidden layers in the network and for introducing new hidden layers (\textit{deepening}). Conversely, Network Morphing proposes alternative techniques for widening and deepening, as well as methods for \textit{kernel size morphing} and \textit{sub-net morphing}.
 
Previous FPTs do not handle residual connections \cite{resnet}, however, they are frequently used to train deep networks faster and more reliably, making them commonplace in the deep learning community. Motivated by this, we propose \R, consisting of two new FPTs \Rwider\ and \Rdeeper, which allow neural networks with residual connections to be morphed.

We believe that the two largest barriers preventing FPTs from being used beyond architecture searches are the complexity overhead of understanding and implementing the transform, alongside a lack of flexibility in the architectures that the FPTs can be applied to. We aim to address these issues by introducing simple to understand and implement FPTs and by facilitating an increase in the number of architectures that FPTs can be applied to. 

Our contributions are as follows: 
(1) proposing \Rwider\ and \Rdeeper, novel FPTs that are compatible with residual connections;
(2) performing a comparative analysis of \R, \N\ and Network Morphism;
(3) introducing a novel evaluation to show that FPTs can be used to train networks to convergence with greater computational efficiency on the Cifar-10 \cite{cifar10}. 

\section{Definitions and notation} \label{sec:notation}

Say tensor $T$ has shape $(\alpha \times\beta \times \gamma)$ to formally denote that $T\in\mathbb{R}^{\alpha\times\beta\times\gamma}$. Also consider a convolutional kernel $W$ and a volume $x$ with shapes $(C_o \times C_i \times k_h \times k_w)$ and $(C_i \times h \times w)$ respectively. Let $W_{ij} * x_k$ be a 2D convolution defined in any standard way, which may include (integer) stride, padding and dilation for example. For notational convenience assume that the spatial dimensions are preserved (i.e. $W_{ij}*x_k$ has shape $(h\times w)$). The arguments made in later sections are valid without the need for this assumption, albeit with a slightly more complex shape analysis.

In convolutional neural networks, a convolution is often considered to operate between 4D and 3D tensors, such that $W*x$ has shape $(C_o \times h\times w)$. This complex \textit{convolution product} can also be defined using matrix notation and the 2D convolution as follows:
\begin{align}
	W*x &\stackrel{\text{def}}{=} 
      	\begin{bmatrix}
          W_{11} & W_{12} & \dots  & W_{1C_i} \\
          W_{21} & W_{22} & \dots  & W_{2C_i} \\
          \vdots & \vdots & \ddots & \vdots \\
          W_{C_o1} & W_{C_o2} & \dots  & W_{C_oC_i}
  		\end{bmatrix}
        *
        \begin{bmatrix}
          x_1 \\
          x_2 \\
          \vdots \\
          x_{C_i} 
  		\end{bmatrix} \nonumber \\
        &= \begin{bmatrix}
          \sum_{m=1}^{C_i} W_{1m}*x_m \\
          \sum_{m=1}^{C_i} W_{2m}*x_m \\
          \vdots \\
          \sum_{m=1}^{C_i} W_{C_om}*x_m 
  		\end{bmatrix}.
\end{align}

$W_{ij}$ and $x_k$ are 2D matrices, with shapes $(k_h\times k_w)$ and $(h\times w)$ respectively. \textit{Block convolutions}, can be considered in a similar fasion to block matrix multiplication. For example, if $A,B,C,D$ have shape $(4,2,5,5)$ and $a,b$ have shape $(2,100,100)$, then we may write:
\begin{align}
    \begin{bmatrix}
        A & B \\
        C & D
  	\end{bmatrix}
    *
    \begin{bmatrix}
      a \\
      b
  	\end{bmatrix} 
  	=
    \begin{bmatrix}
      A*a + B*b \\
      C*a + D*b
  	\end{bmatrix}.
\end{align}

Note that if $k_h=k_w=h=w=1$, then the convolution operation above reduces to matrix multiplication (with some unnecessary additional dimensions). Because of this reduction, we will discuss our function preserving transform only in a convolutional setting, and the corresponding formulation using linear layers follows.

Consider a standard formulation of a convolutional neural network (ignoring residual connections for now). Let the $i^{\text{th}}$ intermediate volume be $\la{x}{i}$ with shape $(\la{C}{i} \times \la{h}{i} \times \la{w}{i})$, let the $i^{\text{th}}$ convolutional kernel be $\la{W}{i}$ with shape $(\la{C}{i} \times \la{C}{i-1} \times \li{k}{i}{h} \times \li{k}{i}{w})$ and corresponding bias $\la{b}{i}$ with shape $(\la{C}{i})$. Consider also a non-linearity $\li{\sigma}{i}{\la{\rho}{i}}:\mathbb{R}^{\la{C}{i} \times \la{h}{i-1} \times \la{w}{i-1}} \rightarrow \mathbb{R}^{\la{C}{i} \times \la{h}{i} \times \la{w}{i}}$ that is applied to the $i^{\text{th}}$ intermediate volume. Parameters $\la{\rho}{i}$ are included in the definition of $\li{\sigma}{i}{\la{\rho}{i}}$ to indicate that there may be some parameters associated with the non-linearity. Given the above definitions, $\la{x}{i+1}$ and $\la{x}{i}$ can be computed using the following equations:
\begin{align} 
    \la{x}{i} =& \li{\sigma}{i}{\la{\rho}{i}} \left( \la{W}{i} * \la{x}{i-1} + \la{b}{i} \right) \label{eqn:layer_i}\\
    \la{x}{i+1} =& \li{\sigma}{i+1}{\la{\rho}{i+1}} \left(\la{W}{i+1} * \la{x}{i} + \la{b}{i+1} \right). \label{eqn:layer_i+1}
\end{align}

In this work \textit{non-linearity} refers to any part of the network other than convolutions and linear (matrix) operations. This includes batch norm \cite{batchnorm}, max pooling and average pooling layers. In our experiments, we compose ReLUs, pooling layers and batch norm into the non-linearity, hence batch norms parameters would be included in $\la{\rho}{i}$.

Additionally, the non-liniarity $\li{\sigma}{i}{\la{\rho}{i}}$ is assumed to operate on each channel independently. That is, for any input $x$ with shape $(\la{C}{i}\times \la{h}{i} \times \la{w}{i})$ we can write:
\begin{align}
     \li{\sigma}{i}{\la{\rho}{i}} \left(
        \begin{bmatrix}
          x_1 \\
          x_2 \\
          \vdots \\
          x_{\la{C}{i}} 
  		\end{bmatrix} \right)
    = 
        \begin{bmatrix}
          \li{\sigma}{i}{\li{\rho}{i}{1}}(x_1) \\
          \li{\sigma}{i}{\li{\rho}{i}{2}}(x_2) \\
          \vdots \\
          \li{\sigma}{i}{\li{\rho}{i}{\la{C}{i}}}(x_{\la{C}{i}}) 
  		\end{bmatrix}.  \label{eqn:nonlinearity_assumption}                
\end{align}

\section{Related work}
We provide an overview of the only two other function preserving transform methods to date (to the best of our knowledge). Then, we consider a few existing applications of these function preserving transformations. 


\subsection{Net2Net}

\N\ introduces two function preserving transforms: \Nwider\ for adding more hidden units into the network layers and \Ndeeper\ for expanding the network with additional hidden layers. To widen layer the $i^{\text{th}}$ dense layer in a network, $\la{h}{i}=\psi(\la{W}{i}\la{h}{i-1})$, from $n$ to $q$ hidden units using \Nwider, the weight matrices $W^{(i)}$ and $W^{(i+1)}$ are replaced with $\la{U}{i}$ and $\la{U}{i+1}$ where:
\begin{align}
U^{(i)}_{j, k} = W^{(i)}_{g(j), k}, && U^{(i+1)}_{h, j} = \dfrac{1}{|\{x \mid g(x) = g(j)\}|}W^{(i+1)}_{h,g(j)}
\end{align}
and where 
\begin{align*}
   g(j) = \begin{cases}
j,& \text{for } j\leq n\\
\text{random sample in } \{1, 2, \dots n\}, & \text{for } n < j \leq q.
\end{cases}
\end{align*}
\Nwider\ duplicates columns from $W^{(i)}$ to widen a hidden layers and obtain $U^{(i)}$ and then adjusts for their replication in $U^{(i+1)}$ to be function preserving. 

\Ndeeper\ can be used to increase the number of hidden layers in a network by replacing a layer $h^{(i)} = \psi(\la{W}{i}\la{h}{i-1})$ with two layers $h^{(i)} = \psi(V^{(i)} \psi(W^{(i)}h^{(i-1)}))$. The weight matrix for the newly added layer $V^{(i)}$ is initialized to the identity matrix. \Ndeeper\ only works with idempotent activation functions $\psi$, where $\psi(I\psi(v))=\psi(v)$. While, ReLU activations satisfy this property, sigmoid and tanh do not.

\subsection{Network morphism}

Network Morphism \cite{netmorph} generalizes \N, and considers a set of \textit{morphisms} between neural networks that can be used to also increase the kernel size in convolutional layers and to introduce more Network In Network style subnetworks \cite{nin}. 

\cite{netmorph} derive their morphisms in a principled way, by introducing the \textit{network morphism equation}, which provides a sufficient condition for new parameters must satisfy in order for the morphism to be function preserving. To widen a network (ignoring non-linearities) their equations reduce down to solving the equation:
\begin{align}
    \la{W}{i} * \la{W}{i-1} &= \begin{bmatrix}
        \la{W}{i} & \la{U}{i}
    \end{bmatrix} * \begin{bmatrix}
        \la{W}{i-1} \\ \la{U}{i-1}
    \end{bmatrix} \nonumber \\
    &= \la{W}{i} * \la{W}{i-1} + \la{U}{i} * \la{U}{i-1}, \label{eqn:netmorph}
\end{align}

for new parameters $\la{U}{i}, \la{U}{i-1}$. They solve this by setting either $\la{U}{i}$ or $\la{U}{i-1}$ to be zero, and we will refer to widening a network in this way as \NM. They also generalize the equations to include non-linearities. 

\cite{netmorph} also define kernel size morphing, which zero pads a convolutional kernel with zeros in the spatial dimension, and subnet morphism, which in essence adapts \Nwider\ to operate on sebnetworks.

\subsection{Architecture search}

Architecture search algorithms can be considered to simultaneously optimize over neural network architectures and their hyper-parameters. This usually involves training many models and comparing their performance, which can be very computationally expensive, requiring hundreds if not thousands of \textit{GPU days}. To optimize over architectures a number of methods have been considered such as using evolutionary strategies or reinforcement learning \cite{as_ne,as_rl}. Moreover, to lower the computational costs \textit{efficient architecture searches} have been developed.


\begin{figure}[ht]
    \begin{center}
   \includegraphics[width=\columnwidth]{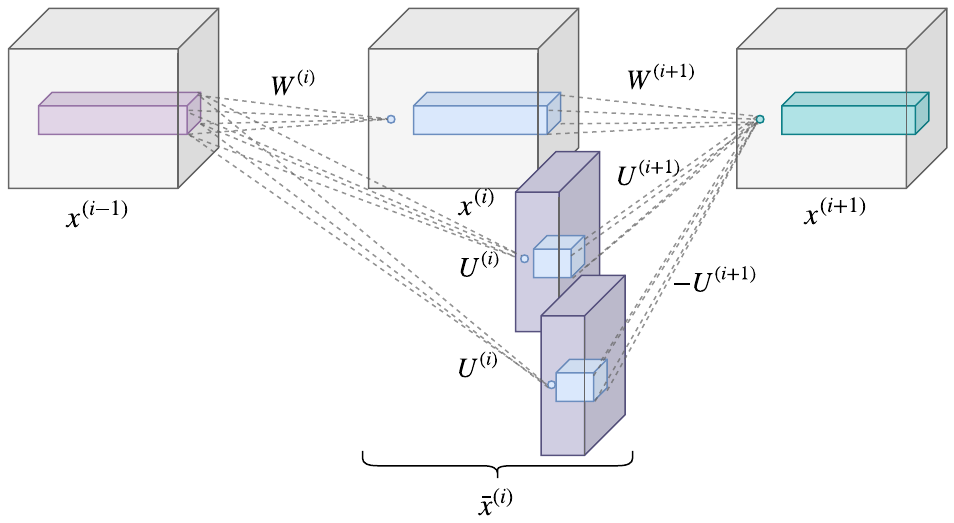}
    \caption{An overview of \Rwider. The purple volumes represent $\li{x}{i}{L}$ and $\li{x}{i}{R}$ in the widened volume $\la{\bar{x}}{i}$. We label the convolution operations with their initializations, and it shows that $\li{x}{i}{L}$ and $\li{x}{i}{R}$ cancel each other out in $\la{x}{i+1}$.}
    \label{fig:r2widerr}
    \end{center}
    \vskip -0.2in
\end{figure}

\N\ and Network Morphism have both been used in \textit{efficient architecture searches}, which aim to reduce the computation time needed to evaluate architectures, thus making architecture search a more approachable technique \cite{n2n_architecture_search,nm_architecture_search}. In parallel to using FPTs, there have been other efforts to making architecture search efficient, such as using a differentiable encoding of architectures and using gradient based algorithms, and using parameter sharing \cite{eas_diff,eas_paramshare}.



\section{\R} \label{sec:r2r}

R2R consists of two FPTs: \Rwider and \Rdeeper, which provide alternative methods for increasing the size of hidden layers and for adding hidden layers in a neural network.  

\subsection{\Rwider} \label{sec:rwider}

To increase the capacity of a network, some intermediate hidden volume $\la{x}{i}$ can be \textit{widened} by increasing its number of channels. We propose \Rwider, a novel way of adding $2E$ channels by padding $\la{x}{i}$ with $\li{x}{i}{L}$ and $\li{x}{i}{R}$, of shapes $(E\times \la{h}{i} \times \la{w}{i})$, thus obtaining a new volume $\la{\bar{x}}{i}$. To assure that the value of $\la{x}{i+1}$ does not change, we initialize new parameters such that $\li{x}{i}{L}=\li{x}{i}{R}$, and that the contributions of $\li{x}{i}{L}$ and $\li{x}{i}{R}$ cancel each other out when computing $\la{x}{i+1}$. This idea is visualized in figure \ref{fig:r2widerr}.

Let $\la{U}{i}$ be an arbitrary tensor with shape $(E\times \la{C}{i-1} \times \la{k_h}{i-1} \times \la{k_w}{i-1})$, let $\la{c}{i}$ be an arbitrary tensor with shape $(E)$ and let $\la{U}{i+1}$ be an arbitrary tensor, with shape $(\la{C}{i+1}\times E \times \la{k_h}{i} \times \la{k_w}{i})$. Additionally, let $\la{\bar{\rho}}{i}$ include all the parameters in $\la{\rho}{i}$ and any additional parameters required, such as batch norm's affine parameters, as discussed previously. We define the new kernels and biases as follows:
\begin{align}
    \la{\bar{W}}{i} &= 
        \begin{bmatrix}
            \la{W}{i} \\
            \la{U}{i} \\
            \la{U}{i}
        \end{bmatrix},\ 
    \la{\bar{b}}{i} =
        \begin{bmatrix}
            \la{b}{i} \\
            \la{c}{i} \\
            \la{c}{i} \\
        \end{bmatrix}, \label{eqn:r2wider_new_params} \\
    \la{\bar{W}}{i+1} &= 
        \begin{bmatrix}
            \la{W}{i+1} &
            \la{U}{i+1} &
            -\la{U}{i+1}
        \end{bmatrix}. 
\end{align}

The intermediate volume $\la{\bar{x}}{i}$ can now be computed: 
\begin{align}
    \la{\bar{x}}{i} =
        \begin{bmatrix}
            \la{x}{i} \\
            \li{x}{i}{L} \\
            \li{x}{i}{R}
        \end{bmatrix}   
    &= \li{\sigma}{i}{\la{\bar{\rho}}{i}}\left( 
            \la{\bar{W}}{i} * \la{x}{i-1} +  \la{\bar{b}}{i}
        \right) \label{eqn:new_layer_i} \\
        & =
        \begin{bmatrix}
            \li{\sigma}{i}{\la{\bar{\rho}}{i}}(\la{W}{i} * \la{x}{i-1} + \la{b}{i})\\
            \li{\sigma}{i}{\la{\bar{\rho}}{i}}(\la{U}{i} * \la{x}{i-1} + \la{c}{i})\\
            \li{\sigma}{i}{\la{\bar{\rho}}{i}}(\la{U}{i} * \la{x}{i-1} + \la{c}{i})
        \end{bmatrix}. \label{eqn:block_operation_layer_i} 
\end{align}

In equation (\ref{eqn:block_operation_layer_i}) it is clear that we have $\li{x}{i}{L}=\li{x}{i}{R}$. Thus,
\begin{align}
    & \la{\bar{W}}{i+1}*\la{\bar{x}}{i} \nonumber \\ 
    &= \la{W}{i+1}*\la{x}{i} + \la{U}{i+1}*\li{x}{i}{L} - \la{U}{i+1}*\li{x}{i}{R} \nonumber \\
    & = \la{W}{i+1}*\la{x}{i}. \label{eqn:conv_eq}
\end{align}

Using equation (\ref{eqn:conv_eq}) to compute the $(i+1)^{\text{th}}$ layer, we obtain: 

\begin{align}
    \la{x}{i+1} 
    =
        \li{\sigma}{i+1}{\la{\rho}{i+1}} \Big( &
        \la{\bar{W}}{i+1} * \la{\bar{x}}{i}
        + \la{b}{i+1} \Big). \label{eqn:new_layer_i+1}
\end{align}

To conclude, the \Rwider\ transformation defines new kernels and biases as in (\ref{eqn:r2wider_new_params}) and replaces equations (\ref{eqn:layer_i}), (\ref{eqn:layer_i+1}) with the equations (\ref{eqn:new_layer_i}) and (\ref{eqn:new_layer_i+1}) in the neural network. We provide a full derivation of \Rwider, in appendix \ref{app:full_derivation}.

\subsection{Residual connections in \Rwider} \label{sec:residual}

\begin{figure}[ht]
    \begin{center}
    \includegraphics[width=\columnwidth]{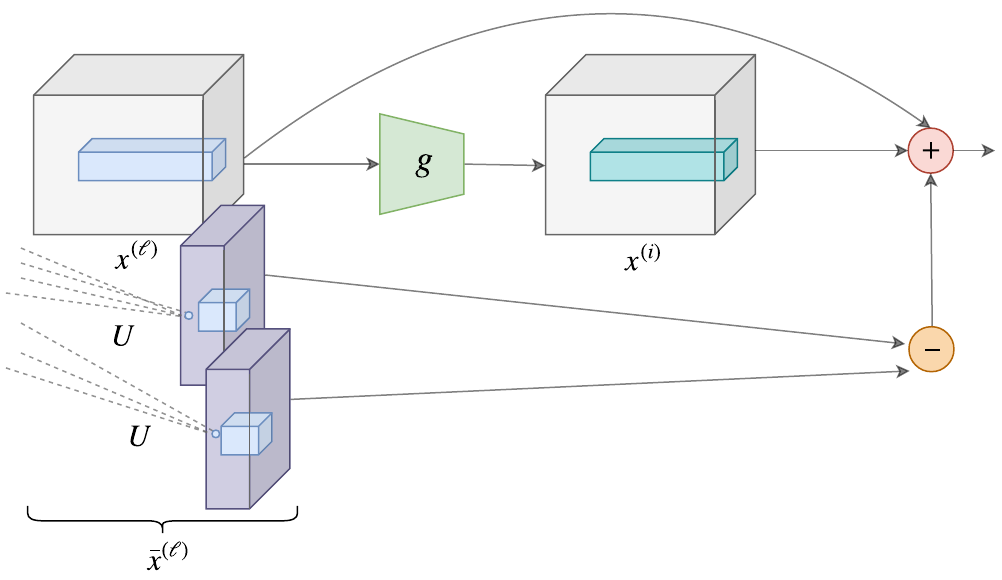}
    \caption{A schematic of how to adapt simple residual connections for the \Rwider\ operation, in the simple case. The purple volumes indicate the new parameters in $\la{\bar{x}}{\ell}$, and $g$ represents the composition of layers $\ell+1$ to $i-1$ in the network.}
    \label{fig:r2widerr_rescon}
    \end{center}
    \vskip -0.2in
\end{figure}

We outline how to adapt residual connections to account for \Rwider\ operations. Figure \ref{fig:r2widerr_rescon} gives a schematic overview of the idea. We consider a simplified case where we do not handle (spatial) shape matching, or \textit{down-sampling}, and only show how to adapt a residual connection for a single application of \Rwider. We provide a more general argument that shows how to handle down-sampling and repeated applications of \Rwider\  in appendix \ref{app:full_derivation}.

Suppose that we have a residual connection from the $\ell^{\text{th}}$ intermediate layer to the $i^{\text{th}}$ layer, which we write as 
\begin{align}
    \la{x}{i} = \li{\ef}{i}{\la{\theta}{i}}\Big(\la{x}{i-1} \Big) 
        + \la{x}{\ell},  \label{eqn:nnlayer_main_2}
\end{align} 
where $\la{\theta}{i} = \{\la{W}{i}, \la{b}{i}, \la{\rho}{i}\}$ and 
\begin{align}
    \la{x}{i} &= \li{\ef}{i}{\la{\theta}{i}}\Big(\la{x}{i-1} \Big) \nonumber \\
    &\stackrel{\text{def}}{=} \li{\sigma}{i}{\la{\rho}{i}} \left( \la{W}{i} * \la{x}{i-1} + \la{b}{i} \right).
\end{align}

If we apply \Rwider\ to $\la{x}{\ell}$, then equation (\ref{eqn:block_operation_layer_i}) gives: 
\begin{align}
    \la{\bar{x}}{\ell} = 
        \begin{bmatrix}
            \la{x}{\ell} \\
            \li{x}{\ell}{L} \\
            \li{x}{\ell}{R} 
        \end{bmatrix} 
    & =  
        \begin{bmatrix}
            \li{\ef}{\ell}{\la{\theta}{\ell}}\Big(\la{x}{\ell-1}\Big) \\
            \li{\ef}{\ell}{\li{\theta}{\ell}{L}}\Big(\la{x}{\ell-1}\Big) \\
            \li{\ef}{\ell}{\li{\theta}{\ell}{R}}\Big(\la{x}{\ell-1}\Big) \\
        \end{bmatrix}, \label{eqn:volume_partition}
\end{align}

with $\li{\theta}{\ell}{L}=\{\li{W}{\ell}{L},\li{b}{\ell}{L},\li{\rho}{\ell}{L}\}=\{\li{W}{\ell}{R},\li{b}{\ell}{R},\li{\rho}{\ell}{R}\}=\li{\theta}{\ell}{R}$ and $\li{x}{\ell}{L}=\li{x}{\ell}{R}$. Therefore, if we let 
\begin{align}
    r\left(\la{\bar{x}}{\ell}\right) = \la{x}{\ell} + \li{x}{\ell}{L} - \li{x}{\ell}{R},
\end{align}

\begin{figure*}[ht]
    \begin{center}
    \includegraphics[width=1.7\columnwidth]{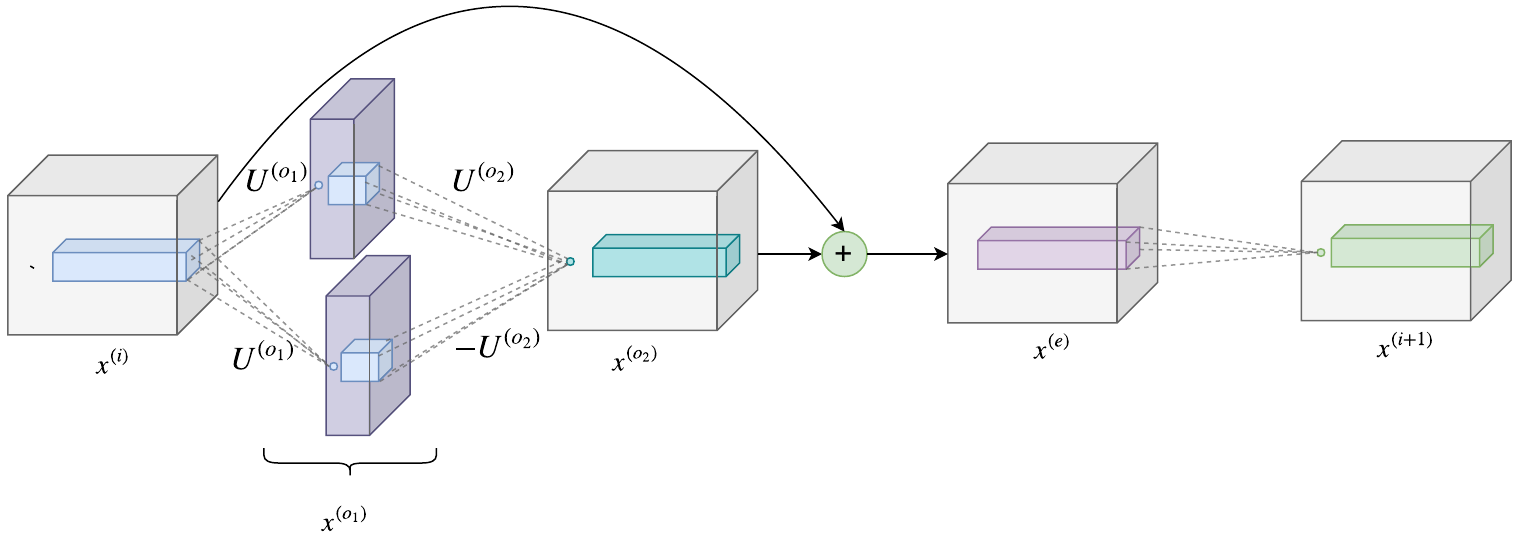}
    \caption{A schematic of the \Rdeeper\ operation. Before \Rdeeper\ is applied, $\la{x}{o_1}$ and $\la{x}{o_2}$ can be ignored, and the network would compute $\la{x}{i+1}$ directly from $\la{x}{i}$. After \Rdeeper\ is applied $\la{U}{o_1}$ is used to make two identical volumes, which cancel each other out in volume $\la{x}{o_2}$, so that $\la{x}{o_2}=0$. The residual connection assures thaen that $\la{x}{e}=\la{x}{i}$.}
    \label{fig:r2deeperr}
    \end{center}
    \vskip -0.2in
\end{figure*}

then we may use 
\begin{align}
    \la{x}{i} = \li{\ef}{i}{\la{\theta}{i}}\Big(\la{x}{i-1} \Big) + r\left(\la{x}{\ell}\right)
\end{align}

in place of equation (\ref{eqn:nnlayer_main_2}), when computing $\la{x}{i}$.

\subsection{Zero initializations} \label{sec:zero_init}

Before we define \Rdeeper, we first consider \textit{zero initializations}, that is, how we can initialize a network such that its output is always zero, while still being able to train it. Thus, so we aim to have as many arbitrary parameters as possible for the zero initialization. Formally, we want to find parameters $\theta_f$, such that $\forall x.\ f(x; \theta_f) = 0$.

Let $g_{\la{\theta}{1:n}}$ be an arbitrary neural network, with parameters $\la{\theta}{1:n}$, and with $n$ layers formed using equations similar to (\ref{eqn:layer_i}) and (\ref{eqn:layer_i+1}). Let $\la{x}{n}=g_{\la{\theta}{1:n}}(\la{x}{0})$. To create a constant zero output, we will add an additional two layers on the output, using the same trick we used to derive \Rwider. 

By letting 
\begin{align}
    \la{W}{o_1} &= \begin{bmatrix}
        \la{U}{o_1} \\
        \la{U}{o_1} \\
    \end{bmatrix}, &
    \la{W}{o_2} &= \begin{bmatrix}
        \la{U}{o_2} &
        -\la{U}{o_2} \\
    \end{bmatrix}, \\
    \la{b}{o_1} &= \begin{bmatrix}
        \la{c}{o_1} \\
        \la{c}{o_1} \\
    \end{bmatrix}, &
    \la{b}{o_2} &= \begin{bmatrix}
        \la{c}{o_2} &
        -\la{c}{o_2} \\
    \end{bmatrix}, 
\end{align}

where $\la{U}{o_1}, \la{U}{o_2}, \la{b}{o_1}, \la{b}{o_2}$ have shapes $(\la{C}{o_1}\times \la{C}{n}\times \la{k_h}{n}\times \la{k_w}{n}), (\la{C}{o_2}\times \la{C}{o_1}\times \la{k_h}{o_1}\times \la{k_w}{o_1}), (\la{C}{o_1})$ and $(\la{C}{o_2})$ respectively, the additional two layers become:
\begin{align} 
    \la{x}{o_1} =& \li{\sigma}{o_1}{\la{\rho}{o_1}} \left( \la{W}{o_1} * \la{x}{n} + \la{b}{o_1} \right), \\
    \la{x}{o_2} =& \li{\sigma}{o_2}{\la{\rho}{o_2}} \left(\la{W}{o_2} * \la{x}{o_1} + \la{b}{o_2} \right). 
\end{align}

Expanding and simplifying, similar to section \ref{sec:rwider}, gives:
\begin{align}
    \la{x}{o_2} =& \li{\sigma}{o_2}{\la{\rho}{o_2}} \left(0 \right).
\end{align}

Therefore, provided we choose $\la{\sigma}{o_2}$ such that $\li{\sigma}{o_2}{{\la{\rho}{o_2}}} \left(0 \right)=0$, then we have a zero initialized network, with output $\la{x}{o_2}$ and $n+2$ layers. See appendix \ref{app:full_derivation} for a full derivation.

\subsection{\Rdeeper} \label{sec:rdeeper}

We propose a novel method for deepening a network using a residual block, initialized to be an identity function. Let $z_\theta$ be a network that is zero initialized in the way described in section \ref{sec:zero_init}, and therefore we have for all $x$ that $z_\theta(x) = 0$. Adding an $x$ to each side then gives 
\begin{align}
    z_\theta(x)+x=x. \label{eqn:identity_deepen}
\end{align} 
Let $\la{x}{e}$ be a new volume we wish to add between the consecutive volumes $\la{x}{i}$ and $\la{x}{i+1}$.
For simplicity assume that $\la{x}{i}$ and $\la{x}{i+1}$ are computed as described in equations (\ref{eqn:layer_i}) and (\ref{eqn:layer_i+1}), however, they be generalized to include residual connections. The layerwise operation now becomes:

\begin{align} 
    \la{x}{i} &= \li{\sigma}{i}{\la{\rho}{i}} \left( \la{W}{i} * \la{x}{i-1} + \la{b}{i} \right),  \\
    \la{x}{e} &= z_\theta\left(\la{x}{i} \right) + \la{x}{i},  \\
    \la{x}{i+1} &= \li{\sigma}{i+1}{\la{\rho}{i+1}} \left(\la{W}{i+1} * \la{x}{e} + \la{b}{i+1} \right). \label{eqn:deepened_out_layer}
\end{align}

From equations (\ref{eqn:identity_deepen}) and (\ref{eqn:identity_deepen}) we must have that $\la{x}{e}=\la{x}{i}$. Substituting $\la{x}{e}$ for $\la{x}{i}$ in equation (\ref{eqn:layer_i+1}), the $(i+1)^{\text{th}}$ layer, gives equation (\ref{eqn:deepened_out_layer}).


\section{A comparison of function preserving transforms}

\begin{table*}[t]
    \label{tab:compare}
    \caption{An overview of the differences between \R, \N\ \cite{net2net} and Network Morphism \cite{netmorph}. We use \begin{small}\textsc{MCD}\end{small} as a shorthand for saying that a function maintains channel dependencies.}
    \vskip 0.15in
    \begin{center}
    \begin{small}
    \begin{sc}
        \begin{tabular}{lccc}
            \toprule
                 & \R & \N & \NM \\
            \midrule
                Activation Functions (Deepen)   & MCD+Fixed Point At Zero & Idempotent & N/A \\
                Activation Functions (Widen)    & MCD & MCD & MCD+Fixed Point At Zero \\ 
                Function Preserving             & $\surd$ & In Excpectation & $\surd$ \\
                Requires noise                  & $\times$ & $\surd$ & $\times$ \\
                Residual Connections            & $\surd$ & $\times$ & $\times$ \\  
                Degrees of Freedom              & Half & None & Half \\ 
                Preserves Existing Parameters   & $\surd$ & $\times$ & $\surd$ \\
                Number of new layers (Deepen)   & $\geq2$ & Any & N/A \\
                Number of new channels (Widen)  & Even & Any & Any \\
            \bottomrule
        \end{tabular}
    \end{sc}
    \end{small}
    \end{center}
    \vskip -0.1in
\end{table*}

This section discusse similarities and differences between each of the FPTs \R, \N\ and \NM, with the aim of providing insight into their use cases. A summary of differences can be seen in table \ref{tab:compare}. 

\subsection{Non-linearities}
Each of the FPTs makes assumptions about the form of non-linearities they can be applied with. This restricts what activation functions can be used. Let $\psi$ denote a non-linearity used in the networks that we are applying the FPTs to. \Rwider, \Nwider\ and \NM\ each need the assumption that the non-linearity \textit{maintains channel dependencies}, as described in equation (\ref{eqn:nonlinearity_assumption}). That is, if $\psi$ is applied to volume $x$ with some $i,j$ such that $i\neq j$ and $x_i=x_j$, then we require $\psi(x)_i=\psi(x)_j$. 

\Rdeeper, as stated in section \ref{sec:zero_init}, requires the last non-linearity in the introduced residual block to have a fixed point at zero $\psi(0)=0$. Whereas, in \Ndeeper\ every non-linearity needs to be idempotent, $\psi(\psi(\cdot))=\psi(\cdot)$. \cite{net2net} incorporate batch normalization by setting its affine parameters to invert the normalization and produce and identity function (for a specific mini-batch).\footnote{Technically it will always be possible to find an input $x$ to a batch norm layer, bn, such that $\text{bn}(x)\neq x$. The closest we can get is finding affine parameters such that $\mathbb{E}[\text{bn}(x)]=x$.}

\cite{netmorph} manage to avoid these problems by introducing \textit{P-activations}, which use a parameter to interpolate between an identity function (which satisfies all the properties discussed above) and any arbitrary non-linearity during training. This method could be applied to any of the FPTs.

\subsection{Residual connections}

\R\ not only allows the incorporation of residual connections in FPTs, but necessarily requires residual connections for \Ndeeper. Our deepening operation is fundamentally different to those considered by \cite{net2net} and \cite{netmorph}, who approach the problem as one of matrix/convolution decomposition, to split one network layer into two. The \Ndeeper\ operation makes use of the matrix decomposition $A=IA$, where $I$ is the identity matrix, whereas \cite{netmorph} consider that more complex, and dense, decompositions could be used. In contrast, \Rdeeper\ introduces a completely new identity function in the middle of a neural network, using the residual connection to provide the identity. \Rdeeper\ allows for a dense initialization of new parameters, and introduces an entire residual block for each application of \Rdeeper, rather than introducing an extra layer for each application.

We note that both \Nwider\ and \NM\ could be adapted to allow for residual connections, similar to section \ref{sec:residual}. For \NM\ they could be introduced in almost identically as for \Rwider, however, for \Nwider it is not straightforward how to do so. 

Although it has been shown that residual connections are not necessary to train deep networks, although they are still widely used to help train networks stably and quickly \cite{inception}.

\subsection{Preservation of parameters}

A useful property for the transformation is whether it preserves already existing parameters, i.e any parameters that existed in the teacher network are not changed in the student network. This property holds for \R\ and \NM, however, it does not for \N. This property allows the teacher and student networks to co-exist with shared parameters, potentially allowing for efficient training of multiple models of different sizes.

\subsection{Visualizing the transforms}

\begin{figure*}[ht]
    \centering
    \begin{tabular}{ccc}
        \includegraphics[scale=2.0]{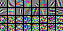} &
        \includegraphics[scale=2.0]{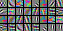} &
        \includegraphics[scale=2.0]{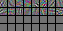} \\
        \includegraphics[scale=2.0]{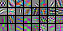} &
        \includegraphics[scale=2.0]{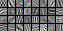} &
        \includegraphics[scale=2.0]{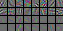} \\
    \end{tabular}
    \caption{Visualization of weights from a $7\times 7$ convolution layer trained on Cifar-10. From left to right we have \R, \N\ and \NM\ schemes. On the top row we have visualizations immediately after the FPT operation, and the bottom row shows the filters after they have all been trained to convergence. }
    \label{fig:viz}
\end{figure*}

To help demonstrate the differences between \Rwider\, \Nwider\ and \NM\ we visualize the weights in the first layer of a a three layer convolutions network, initially containing 16 filters. We train the network on the Cifar-10 classification task \cite{cifar10} until convergence. Then we apply \Rwider, \Nwider\ or \NM, to add another 16 filters to the first layer, giving a total of 32. After widening we train again until convergence and compare the weights to when the network was widened. The visualizations can be seen in figure \ref{fig:viz}.  

Because in \Rwider\ and \NM\ there is freedom to choose how about half of the new parameters are initialized, we tended to see some more randomness in the training of the new filters and the filters they converge to. In contrast, in \Nwider\ there is no freedom in the parameter initialization, and we found the \textit{orientation} of the new filters often did not change from they initial set-up. However, the colors being detected  typically changed. As an example, if a new filter from \Nwider\ was initialized as a pink/green vertical edge detector, after further training it may become a blue/orange edge detector. 

Finally the visualizations for \Rwider\ indicate that the new 16 filters were able to learn different filter orientations and that their symmetry was broken without having to add more noise  (as required in \N).  However, we found that the filters learned by \Rwider\ were often noisy themselves. With \NM\ we found that the new filters tended to be quite noisy and/or contain relatively small weights, where the latter is depicted in figure \ref{fig:viz}. We think that this noise in \Rwider\ and the small-weights in \NM\ could be a result of a faint \textit{training signal}, after the teacher network has already learned to classify many examples in the training set correctly.


For all three FPTs, the first 16 filters learned tend to still be present in the larger set of 32 filters after widening and training to convergence again. However, in \Rwider\ initialising the weights with a larger scale compared to the existing filters results in loosing the first 16 filters. 

 

\section{Experiments}


We consider training ResNet-18 networks on Cifar-10 to compare the performance of the FPTs. In appendix \ref{app:hyperparam} we define the ResNetCifar-10(r) and ResNetCifar-18(r) architectures in more detail, where the standard ResNet architectures \cite{resnet} have been adapted for the smaller images size in Cifar-10. r denotes that we use r times as many filters in every convolutional layer, We chose r to ensure that the performance on Cifar-10 is limited in the teacher network, so that the student network has something to improve upon.


In all of our experiments we initialize the free parameters with standard deviation equal to the numerical standard deviation of the existing weights in the kernel for widening and the standard deviation of the weights in kernel prior when deepening. Specifically, if we wanted to initialize new weights with numerical variance $s^2$ then we sampled from the uniform distribution $U(-\frac{s}{\sqrt{3}},\frac{s}{\sqrt{3}})$. 

All training hyper-parameters are detailed in appendix \ref{app:hyperparam}.

\subsection{Network convergence tests} \label{sec:net_converge_tests}

In line with prior work we perform tests similar to \cite{net2net}, and we compare the converged accuracies when using different FPTs. To test the widening operators we compare performance of ResNetCifar-18($2^{-3}$) networks on Cifar-10. To begin with, a teacher network, with $\sqrt{2}$ fewer channels (i.e. a ResNetCifar-18($2^{-3.5})$ network) is trained to convergence. This network is then used as a teacher network for each of \Rwider, \Nwider\ and \NM, and the respective student networks are trained to convergence. Additionally, we also compare with a student using a random padding, and with a ResNetCifar-18($2^{-3}$) initialized completely randomly. 

\begin{figure}[h]
    \centering
    \includegraphics[width=\columnwidth]{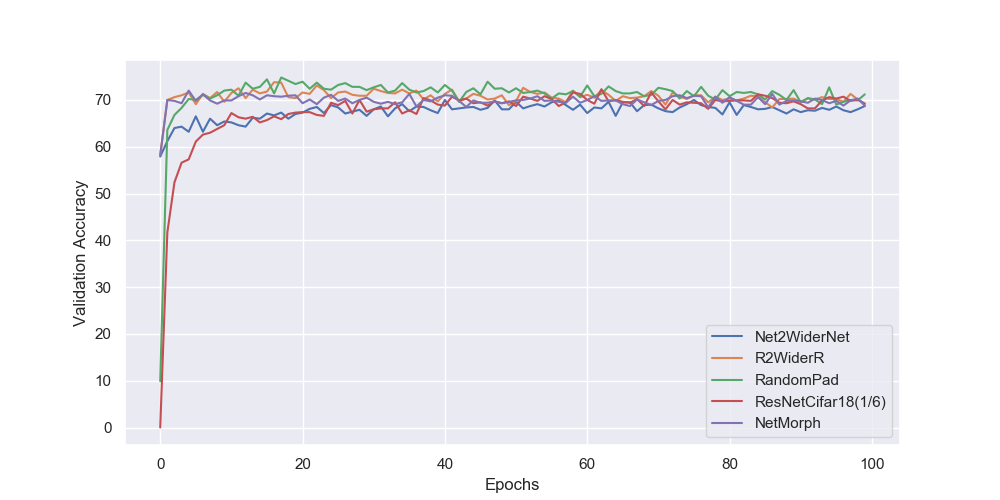}
    \caption{Validation curves comparing a student network using each of \Nwider, \Rwider, \NM, with baselines of random padding and training the ResNet ``from scratch''. }
    \label{fig:n2wn_val_results}
\end{figure}

We found that all three FPTs are competitive and converged to at least as good score as the randomly initialized networks. To apply \Nwider\ we had to remove the residual connections from ResNetCifar-18($2^{-3}$). We found that this particular network, similar to \cite{resnet}, could not reach the same performance as the one with residual connections; thus, we believe that it would be unfair, given this data, to draw any direct conclusions on the performance of \N\ versus \R. Validation curves are shown in figure \ref{fig:n2wn_val_results} and final validation accuracies are in table \ref{tab:n2n_val_results_table}.

\begin{figure}[H]
    \centering
    \includegraphics[width=\columnwidth]{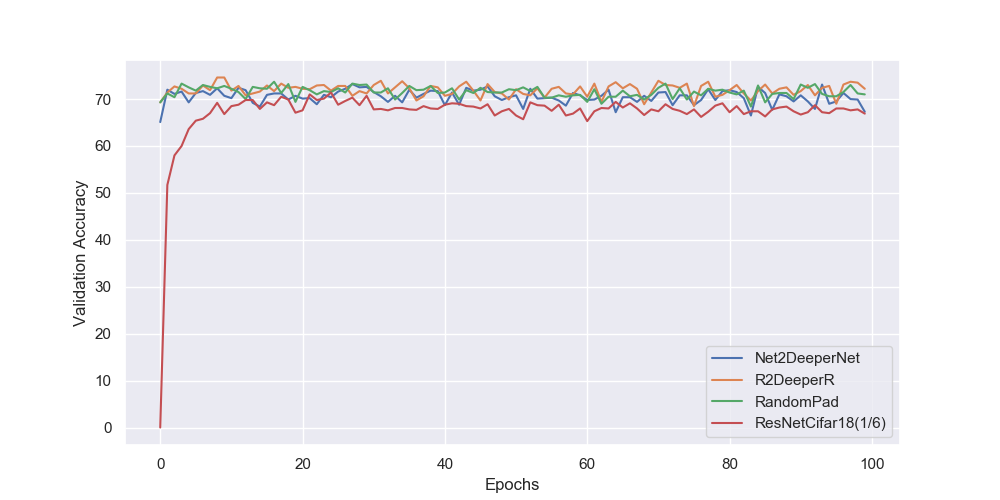} 
    \caption{Validation curves comparing a student network using each of \Ndeeper, \Rdeeper, \NM, with baselines of random padding and training the ResNet ``from scratch''. }
    \label{fig:n2dn_val_results}
\end{figure}

To test deepening operators we compare the performance of ResNetCifar-18($2^{-3}$) networks on Cifar-10. This time we use a ResNetCifar-10($2^{-4}$) for the teacher network, and then use either \Rdeeper\ or \Ndeeper to add eight layers. We also compare with random padding and a randomly initialized ResNetCifar-18($2^{-3}$). Validation curves are in figure \ref{fig:n2dn_val_results}, and final accuracies in table \ref{tab:n2n_val_results_table}.

\begin{table}[ht]
    \caption{Final validation scores for student networks at the \textit{end} of training. All networks reach similar performance.}
    \vskip 0.15in
    \begin{center}
    \begin{small}
    \begin{sc}
        \begin{tabular}{lc}
            \toprule
                 & Validation Accuracy \\
            \midrule
                ResNetCifar($2^{-3}$)       & 69.9\% \\
                \Rwider\ student            & 69.9\%  \\ 
                \Nwider\ student            & 69.6\%  \\ 
                \NM\ student                & 69.8\%  \\ 
                RandomPadWiden student      & 70.2\%  \\
                \Rdeeper\ student           & 72.1\%  \\ 
                \Ndeeper\ student           &  70.5\%  \\ 
                RandomPadDeepen student     & 71.5\%  \\
            \bottomrule
        \end{tabular}
    \end{sc}
    \end{small}
    \end{center}
    \label{tab:n2n_val_results_table}
    \vskip -0.1in
\end{table}

\begin{figure}[ht]
    \centering
    \includegraphics[width=0.9\columnwidth]{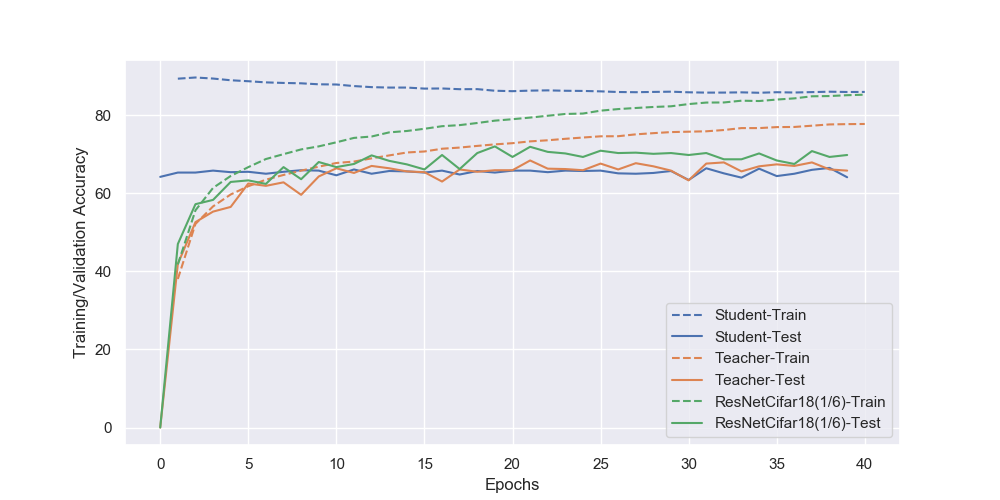} 
    \caption{In this example we training curves where the teacher network (a ResNetCifar-18($2^{-3.5}$)) was allowed to overfit. The student network (ResNetCifar-18($2^{-3}$) is unable to make any additional performance gains, and a randomly initialized ResNetCifar-18($2^{-3}$) is able to reach a better converged performance.}
    \label{fig:overfitting_results}
    \vskip -0.2in
\end{figure}


We also observed that if the teacher network overfits during training, then the student has minimal improvement over the teacher, as can be seen in figure \ref{fig:overfitting_results}. This can be explained by the weak training signal due to the teacher having overfitted, and it therefore struggles to improve on the initial performance. Due to this phenomenon we conclude that regularization to preventing overfitting is essential for a student network to perform better than the teacher.

\begin{figure}[h]
    \centering
    \begin{tabular}{c}
        \includegraphics[width=\columnwidth]{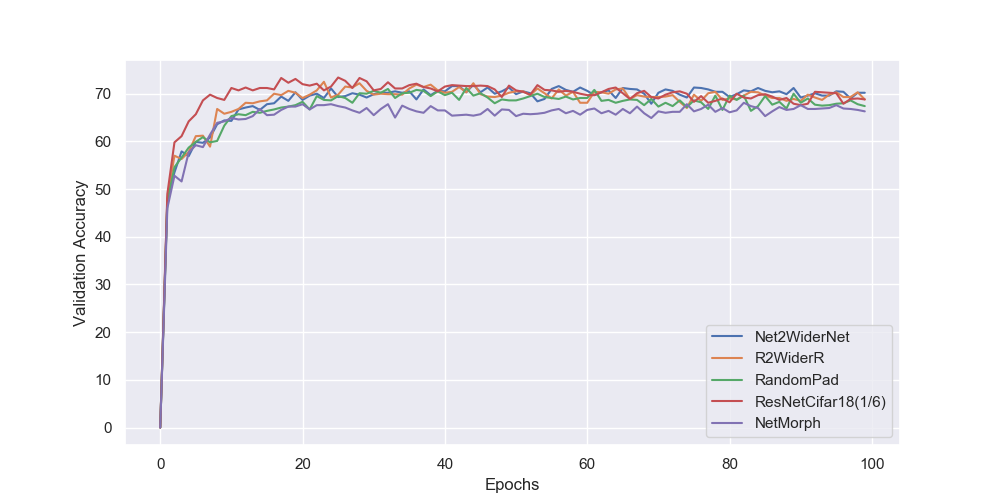} \\
        \includegraphics[width=\columnwidth]{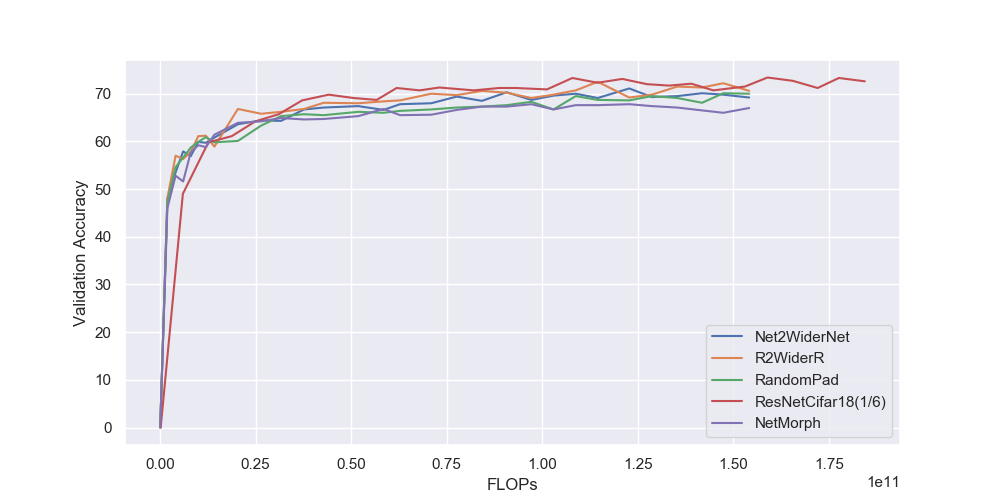} 
    \end{tabular}
    \caption{Validation curves comparing a network using each of \Ndeeper, \Rdeeper, to deepen at 25 epochs. We compare with baselines of random padding and training the ResNet ``from scratch''. Similar results for widening operations are omitted for space. \textit{Top}: Plots with respect to the number of epochs passed (validation score with respect to sample complexity). \textit{Bottom}: Plots with respect to the number of FLOPs used (validation score with respect to computational complexity, only 30 epochs shown). }
    \label{fig:r2r_results}
    \vskip -0.1in
\end{figure}

\subsection{Faster training tests} \label{sec:faster_convergence}

We also consider if FPTs can be used to train a network to convergence faster than initializing the network from scratch, while still achieving the same performance.In these experiments we train a ResNetCifar-18($2^{-3.5}$) or ResNetCifar-10($2^{-3}$) teacher network and then appropriately widening or deepening the network using \R, \N\ or Network Morphism in the middle of training. Each network is then trained until it converges.

In figure \ref{fig:r2r_results} we see that training is slower with respect to the number of training updates. However, when we consider the actual number of floating point operations (FLOPs) we find that using the FPTs was faster. This suggests that FPTs can be used in similar training procedures to trade off between computational and sample complexities of training.


These experiments also indicate that we need to be careful when initializing new parameters in \R. We observed that if the new weights were initialized with large values with respect to existing weights in the network it lead to instability in the training, and a drop in performance immediately after the widen. We illustrate this case in the Appendix \ref{apx:bad_weight_init}.

The results obtained in this section indicate that FPTs can be used to train our ResNetCifar-18 networks with a lower computational cost and similar performance.


\section{Conclusion}

In this work we have introduced new FPTs: \Rwider\ and \Rdeeper, that incorporate residual connections into the theory of FPTs. We derived how they preserve the function represented by the networks, despite altering the architecture. We then provided an in depth discussion on the differences, similarities and use-cases of \R, \N\ and \NM. Experiments conducted on Cifar-10 demonstrated that all three FPT schemes have similar performance, thereby allowing a wide range of neural network architectures to have FPTs applied to them. Finally, we also demonstrated that FPTs can be used to trade off between sample complexity and computational complexity of training.


\bibliographystyle{icml2019}
\bibliography{main.bib}

\newpage
\onecolumn
\appendix

\section{Code} \label{app:A}

Code implemented in PyTorch (whichever version was current at the start of 2019): \href{https://github.com/MWPainter/Deep-Neuroevolution-With-SharedWeights---Ensembling-A-Better-Solution}{https://github.com/MWPainter/Deep-Neuroevolution-With-SharedWeights---Ensembling-A-Better-Solution}.

I spent a bit of time after the ICML submission trying to address the concerns around networks achieving closer to state-of-the-art performance and having results with larger datasets than Cifar-10.

\section{Some Clearer Plots} \label{app:B}

\begin{figure}[h]
    \centering
    \begin{tabular}{cc}
        \includegraphics[width=0.49\columnwidth]{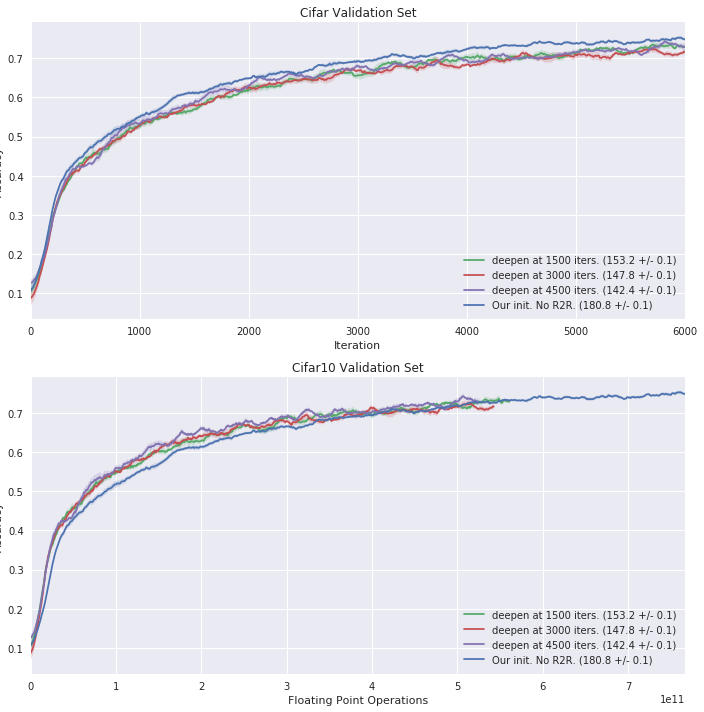} & \includegraphics[width=0.49\columnwidth]{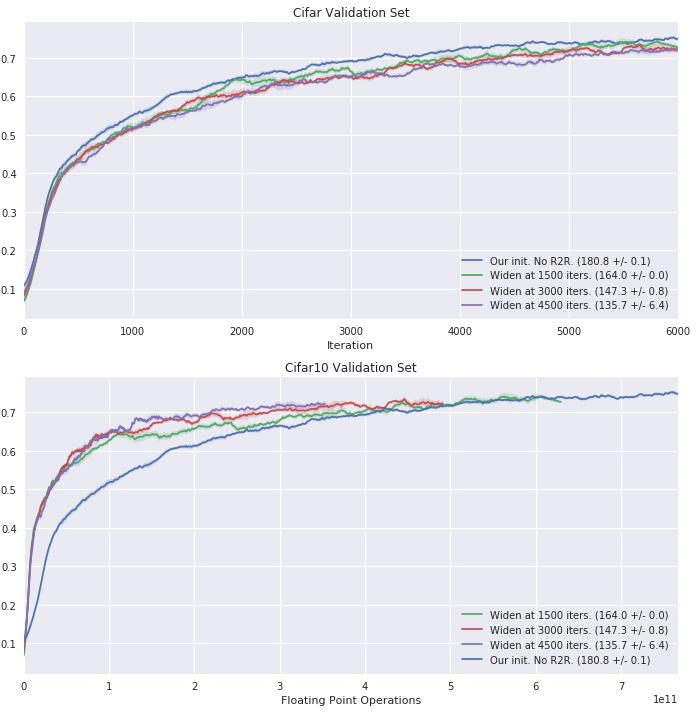} 
    \end{tabular}
    \caption{Plots from the original CS231n project report, similar to Figure 8. Left column shows results for \Rdeeper and right column shows results for \Rwider. The top row shows training curves with respect to the number of network updates, and the bottom row shows the same curves with respect to the number of floating point operations performed (i.e. the amount of GPU compute required). The numbers in the legend (such as $153.2\pm 0.1$) indicate the total time that it took to train that network (including any overheads from loading libraries etc)}
    \label{fig:r2r_results_project}
\end{figure}

\section{A full derivation of \R} \label{app:full_derivation}

In this section we provide an extended descriptions of the steps we went through to derive \R\ so that the padding produces a function preserving transform. In particular we provide a more principled approach by finding equations that must hold for the transformation/padding to be function preserving and then solving them. Solutions can then be found by observation, yielding the methods presented in section \ref{sec:r2r}.

Additional note for arxiv version of this paper: I recall having some maths (that I never got around to writing up) that suggested that the symmetry introduced in the zero-initializations should be broken after some number of backups (if I remember correctly it did rely on either the input/output data not being symmetric in some sense that I cannot recall), which can be observed in Figure 4. However, in retrospect, some of the new filters are still quite symmetrical, so adding a small amount of noise to any new parameters may be helpful for symmetry breaking.

\subsection{\Rwider}

Recall equations (\ref{eqn:layer_i}) and (\ref{eqn:layer_i+1}) which we used to define the operation of a neural network. Suppose that we want to widen $\la{x}{i}$ by padding with $\li{x}{i}{L}$ and $\li{x}{i}{R}$, of shapes $(E\times \la{h}{i} \times \la{w}{i})$, to produce a new volume $\la{\bar{x}}{i}$. Let $\li{W}{i}{L}, \li{W}{i}{R}$ with shape $(E\times \la{C}{i-1} \times \la{k_h}{i-1} \times \la{k_w}{i-1})$, $\li{W}{i+1}{L}, \li{W}{i+1}{R}$ with shape $(\la{C}{i+1}\times E \times \la{k_h}{i} \times \la{k_w}{i})$, and $\li{b}{i}{L}, \li{b}{i}{R}$ with shape $(E)$ be the new parameters introduced.
Also, we introduce new parameters $\li{\rho}{i}{L},\li{\rho}{i}{R}$ for additional channels required in the non-linearity $\la{\sigma}{i}$, and let $\la{\bar{\rho}}{i} = \la{\rho}{i} \cup \li{\rho}{i}{L} \cup \li{\rho}{i}{R}$.

Given this padding, the $i^{\text{th}}$ intermediate volume in the neural network is now $\la{\bar{x}}{i}$ and has shape $((\la{C}{i}+2E) \times \la{h}{i} \times \la{w}{i})$. Let the new $(i+1)^{\text{th}}$ intermediate volume be denoted $\la{\bar{x}}{i+1}$, which has the same shape as $\la{x}{i+1}$ of $(\la{C}{i+1}\times E \times \la{k_h}{i} \times \la{k_w}{i})$. 

Equations (\ref{eqn:before_widen1}) and (\ref{eqn:before_widen2}) below recall the operation of the neural network before the padding, and equations (\ref{eqn:after_widen1}) and (\ref{eqn:after_widen2}) show the new operation after the padding.

\begin{align} 
    \la{x}{i} &= \li{\sigma}{i}{\la{\rho}{i}}\left( 
        \la{W}{i} * \la{x}{i-1} + \la{b}{i} \right) \label{eqn:before_widen1}
    \\
    \la{x}{i+1} &= \li{\sigma}{i+1}{\la{\rho}{i+1}}\left( 
        \la{W}{i+1} * \la{x}{i} + \la{b}{i} \right) \label{eqn:before_widen2}
    \\
        \begin{bmatrix}
            \la{x}{i} \\
            \li{x}{i}{L} \\
            \li{x}{i}{R}
        \end{bmatrix} =
    \la{\bar{x}}{i} 
    &=
    \li{\sigma}{i}{\la{\bar{\rho}}{i}}\left( 
        \begin{bmatrix}
            \la{W}{i} \\
            \li{W}{i}{L} \\
            \li{W}{i}{R}
        \end{bmatrix} * \la{x}{i-1} + 
        \begin{bmatrix}
            \la{b}{i} \\
            \li{b}{i}{L} \\
            \li{b}{i}{R} \\
        \end{bmatrix}
        \right) \label{eqn:after_widen1}
    \\
    \la{\bar{x}}{i+1} 
    &= 
    \li{\sigma}{i}{\la{\bar{\rho}}{i+1}}\left( 
        \begin{bmatrix}
            \la{W}{i+1} &
            \li{W}{i+1}{L} &
            \li{W}{i+1}{R}
        \end{bmatrix} * \la{\bar{x}}{i} + 
            \la{b}{i+1} 
        \right) \label{eqn:after_widen2}
\end{align}

By expanding $\la{\bar{x}}{i}$ in equation (\ref{eqn:after_widen2}) we obtain
\begin{align}
    \la{\bar{x}}{i+1} &= 
        \li{\sigma}{i}{\la{\bar{\rho}}{i+1}}\left( 
            \begin{bmatrix}
                \la{W}{i+1} &
                \li{W}{i+1}{L} &
                \li{W}{i+1}{R}
            \end{bmatrix} * \begin{bmatrix}
                \la{x}{i} \\
                \li{x}{i}{L} \\
                \li{x}{i}{R}
            \end{bmatrix} 
            + \la{b}{i+1} 
            \right) \\
        &= 
        \li{\sigma}{i}{\la{\bar{\rho}}{i+1}}\left( 
                \la{W}{i+1} * \la{x}{i} +
                \li{W}{i+1}{L} * \li{x}{i}{L} +
                \li{W}{i+1}{R} * \li{x}{i}{R} 
            + \la{b}{i+1} 
            \right).
\end{align}

So it is sufficient that $\li{W}{i+1}{L}=-\li{W}{i+1}{R}$ and $\li{x}{i}{L}=\li{x}{i}{R}$ for $\la{x}{i+1}=\la{\bar{x}}{i+1}$. Then, recalling equation (\ref{eqn:nonlinearity_assumption}), our assumption about the form of the non-lineariy, we have
\begin{align}
        \begin{bmatrix}
            \la{x}{i} \\
            \li{x}{i}{L} \\
            \li{x}{i}{R}
        \end{bmatrix}   
    = \li{\sigma}{i}{\la{\bar{\rho}}{i}}\left( 
        \begin{bmatrix}
            \la{W}{i} \\
            \li{W}{i}{L} \\
            \li{W}{i}{R}
        \end{bmatrix} * \la{x}{i-1} + 
        \begin{bmatrix}
            \la{b}{i} \\
            \li{b}{i}{L} \\
            \li{b}{i}{R} \\
        \end{bmatrix}
        \right) 
         =
        \begin{bmatrix}
            \li{\sigma}{i}{\la{\rho}{i}}(\la{W}{i} * \la{x}{i-1} + \la{b}{i})\\
            \li{\sigma}{i}{\li{\rho}{i}{L}}(\li{W}{i}{L} * \la{x}{i-1} + \li{b}{i}{L})\\
            \li{\sigma}{i}{\li{\rho}{i}{R}}(\li{W}{i}{R} * \la{x}{i-1} + \li{b}{i}{R})
        \end{bmatrix}. \label{eqn:layer_i_with_non_linear_assumption}
\end{align}

And so if we set $\li{W}{i}{L}=\li{W}{i}{R}$, $\li{b}{i}{L}=\li{b}{i}{R}$ and $\li{\rho}{i}{L}=\li{\rho}{i}{R}$, then we have $\li{x}{i}{L}=\li{x}{i}{R}$. 

In conclusion, if we have $\la{\sigma}{i}$ that satisfies the assumption in equation (\ref{eqn:nonlinearity_assumption}), and set $\li{W}{i}{L}=\li{W}{i}{R}, \li{b}{i}{L}=\li{b}{i}{R}, \li{\rho}{i}{L}=\li{\rho}{i}{R}, \li{W}{i+1}{L}=-\li{W}{i+1}{R}$ (where we can choose $\li{W}{i}{R}, \li{\rho}{i}{R}, \li{W}{i+1}{R}, \li{b}{i}{R}$ arbitrarily), then the transform is function preserving. We also note that \Rwider\ can be considered a non-trivial solution to the network morphism equations defined by \netmorphauth\ \yrcite{netmorph}.

\subsection{Residual connections} 

In section \ref{sec:residual} we outlined how residual connections can be adapted for \Rwider, however, we did not cover how to deal with either widening a volume multiple times or how to handle down-sampling. In this section we generalize the argument to handle both of these cases.

For convenience let $\la{\theta}{i} = \{\la{W}{i}, \la{b}{i}, \la{\rho}{i}\}$ and define $\ef$ as a shorthand for equation (\ref{eqn:before_widen1}) as follows:
\begin{align}
    \la{x}{i} = \li{\ef}{i}{\la{\theta}{i}}\Big(\la{x}{i-1} \Big) 
    &\stackrel{\text{def}}{=} \li{\sigma}{i}{\la{\rho}{i}} \left( \la{W}{i} * \la{x}{i-1} + \la{b}{i} \right). \label{eqn:nnlayer_1} 
\end{align}

Consider if \Rwider\ is applied to an intermediate volume $\la{x}{\ell}$, producing the two equal volumes $\li{x}{\ell}{L}$ and $\li{x}{\ell}{R}$. If $\la{x}{\ell}$ was used as part of a residual connection, then we need to take into account how to handle $\li{x}{\ell}{L}$ and $\li{x}{\ell}{R}$ over that residual connection. We can re-write equation (\ref{eqn:layer_i_with_non_linear_assumption}) as:
\begin{align}
    \la{\bar{x}}{\ell} = 
        \begin{bmatrix}
            \la{x}{\ell} \\
            \li{x}{\ell}{L} \\
            \li{x}{\ell}{R} 
        \end{bmatrix} 
    & =  
        \begin{bmatrix}
            \li{\ef}{\ell}{\la{\theta}{\ell}}\Big(\la{x}{\ell-1}\Big) \\
            \li{\ef}{\ell}{\li{\theta}{\ell}{L}}\Big(\la{x}{\ell-1}\Big) \\
            \li{\ef}{\ell}{\li{\theta}{\ell}{R}}\Big(\la{x}{\ell-1}\Big) \\
        \end{bmatrix}, \label{eqn:volume_partition}
\end{align}

Let $\li{r}{i}{\la{\phi}{i}}$, with parameters $\la{\phi}{i}$, denote a function that \textit{reshapes} a tensor to have shape $(\la{C}{i}\times \la{h}{i}\times \la{w}{i})$, where the input tensor shape can depend on $\la{\phi}{i}$. The purpose of $\la{r}{i}$ is a parameterized function that is used over the residual connection. To give some examples of what $\la{r}{i}$ could be, if we have $(\la{C}{i},\la{h}{i},\la{w}{i})=(\la{C}{\ell},\la{h}{\ell},\la{w}{\ell})$ then $\la{r}{i}$ would typically be an identity function, if $\la{C}{\ell} < \la{C}{\ell}$ the $\la{r}{i}$ may perform a zero padding and if either $\la{C}{\ell} > \la{C}{i}$, $\la{h}{\ell} > \la{h}{i}$ or $\la{w}{\ell} > \la{w}{i}$ then $\la{r}{i}$ must be some function that downsamples, typically a convolution (and why parameters are needed in $\la{r}{i}$ for this to be a general argument).

Now, consider  a residual connection from the $\ell^{\text{th}}$ intermediate volume to the $i^{\text{th}}$ intermediate volume, formally written as:
\begin{align}
    \la{x}{i} = \li{\ef}{i}{\la{\theta}{i}}\Big(\la{x}{i-1} \Big) 
        + \li{r}{i}{\la{\phi}{i}}\left(\la{x}{\ell}\right).  \label{eqn:nnlayer_2}
\end{align} 

Let $\li{\phi}{i}{L}$ be some arbitrary parameters, such that $\li{r}{i}{\li{\phi}{i}{L}}$ takes tensors with the same shape as $\li{x}{\ell}{L}$. Define $\li{\phi}{i}{R}$ similarly for $\li{x}{i}{R}$. We can now define $\la{\bar{r}}{i}$ to be used as a new residual connection from $\la{\bar{x}}{\ell}$:
\begin{align}
    \la{\bar{r}}{i} \left(\la{\bar{x}}{\ell} \right)
        = \li{r}{i}{\la{\phi}{i}}\left(\la{x}{\ell}\right)
        + \li{r}{i}{\li{\phi}{i}{L}}\left(\li{x}{\ell}{L}\right)
        - \li{r}{i}{\li{\phi}{i}{R}}\left(\li{x}{\ell}{R}\right).
\end{align}

As in \Rwider\ we have $\li{\theta}{i}{L}=\li{\theta}{i}{R}$ in equation (\ref{eqn:volume_partition}), we have $\li{x}{\ell}{L}=\li{x}{\ell}{R}$, and so by setting $\li{\phi}{i}{L}=\li{\phi}{i}{R}$ initially, we have $\la{\bar{r}}{i} (\la{\bar{x}}{\ell}) = \li{r}{i}{\la{\phi}{i}}(\la{x}{\ell})$. Which gives
\begin{align}
    \la{x}{i} 
    &= \li{\ef}{i}{\la{\theta}{i}}\Big(\la{x}{i-1} \Big) +          
            \la{r}{i}\Big(\la{x}{\ell}\Big) \\
    &=  \li{\ef}{i}{\la{\theta}{i}}\Big(\la{x}{i-1} \Big) +          
            \la{\bar{r}}{i}\Big(\la{\bar{x}}{\ell}\Big).
\end{align}

and shows that $\la{\bar{r}}{i}$ can be used in place of the 

Finally, we return to what we added in this section over section \ref{sec:residual}. Our more general argument included non-trivial residual connections, that can include arbitrary parametereized functions and handle down-sampling over the connection. Moreover, because the residual connection includes an arbitrary function, it allows for a recursive application of these rules, and therefore can handle multiple applications of \Rwider over a residual connection.

\subsection{Zero initializations}

Here we provide a full derivation of our zero initializations that we use in \Rdeeper. Let $g_{\la{\theta}{1:n}}$ be an arbitrary neural network, with parameters $\la{\theta}{1:n}$ and let $\la{x}{n}=g_{\la{\theta}{1:n}}(\la{x}{0})$ be the output of this network. We will add two additional laters on the output, $\la{x}{o_1}$ and $\la{x}{o_2}$, with shapes $(\la{C}{o_1}\times \la{h}{o_1} \times \la{w}{o_1})$ and $(\la{C}{o_2}\times \la{h}{o_2} \times \la{w}{o_2})$ respectively.

Assuming that $\la{C}{o_1}$ and $\la{C}{o_2}$ are even, let $\li{W}{o_1}{L},\li{W}{o_1}{R}$ have shape $(\la{C}{o_1}/2\times \la{C}{n}\times \la{h}{n} \times \la{w}{n})$ and let $\li{W}{o_2}{L},\li{W}{o_2}{R}$ have shape $(\la{C}{o_2}\times \la{C}{o_1}/2\times \la{h}{n} \times \la{w}{n})$. Also let $\li{b}{o_1}{L},\li{b}{o_2}{R}$ have shape $(\la{C}{o_1}/2)$ and let $\la{b}{o_2}$ have shape $(\la{C}{o_2})$. Additionally define $\la{\rho}{o_1}=\{\li{\rho}{o_1}{L}, \li{\rho}{o_1}{R}\}$ and $\la{\rho}{o_2}=\{\li{\rho}{o_2}{L}, \li{\rho}{o_2}{R}\}$. Now, define new parameters for the network:
\begin{align}
    \la{W}{o_1} &= \begin{bmatrix}
        \li{W}{o_1}{L} \\
        \li{W}{o_1}{R} \\
    \end{bmatrix}, \\
    \la{W}{o_2} &= \begin{bmatrix}
        \li{W}{o_2}{L} &
        \li{W}{o_2}{R} \\
    \end{bmatrix}, \\
    \la{b}{o_1} &= \begin{bmatrix}
        \li{b}{o_1}{L} \\
        \li{b}{o_1}{R} \\
    \end{bmatrix}.
\end{align}

The computation for new layers $o_1$ and $o_2$ is then:
\begin{align} 
    \la{x}{o_1} =& \li{\sigma}{o_1}{\la{\rho}{o_1}} \left( \la{W}{o_1} * \la{x}{n} + \la{b}{o_1} \right), \\
    \la{x}{o_2} =& \li{\sigma}{o_2}{\la{\rho}{o_2}} \left(\la{W}{o_2} * \la{x}{o_1} + \la{b}{o_2} \right). 
\end{align}

Expanding the definitions of the new params and multiplying out we get:
\begin{align}
    &\la{x}{o_2} = 
        \li{\sigma}{o_2}{\la{\rho}{o_2}} \Big(
            \li{W}{o_1}{L} * \li{\sigma}{o_1}{\li{\rho}{o_1}{L}}\left(\li{W}{o_1}{L} * \la{x}{n} + \li{b}{o_1}{L}\right) + \li{W}{o_2}{R} * \li{\sigma}{n+1}{\li{\rho}{n+1}{R}}\left(\li{W}{n+1}{R} * \la{x}{n} + + \li{b}{o_1}{R}\right)  
        + \la{b}{o_2} \Big). 
        \label{eqn:expanded_zero_init_deriv}
\end{align} 

If we set $\li{W}{o_1}{L}=\li{W}{o_1}{R},\li{b}{o_1}{L}=\li{b}{o_1}{R},\li{W}{o_2}{L}=-\li{W}{o_2}{R}$ and $\li{\rho}{o_1}{L}=\li{\rho}{o_1}{R}$, where $\li{W}{o_1}{R},\li{b}{o_1}{R},\li{W}{o_2}{R}$ and $\li{\rho}{o_1}{R}$ can be arbitrary, then  equation (\ref{eqn:expanded_zero_init_deriv}) above reduces to
\begin{align}
    &\la{x}{o_2} = 
        \li{\sigma}{o_2}{\la{\rho}{o_2}} \Big( 
            \la{b}{o_2} \Big). 
\end{align}

As $\la{x}{o_2}$ will be the output volume, if $\li{\sigma}{o_2}{\la{\rho}{o_2}}(\la{b}{o_2})=0$, then we have found a zero initialization. In particular, we use ReLU activations in our experiments, which satisfy $\li{\sigma}{o_2}{\la{\rho}{o_2}}(0)=0$, in which case we set $\la{b}{o_2}=0$ as well. 

We note that there is a restriction on the output non-linearity, which we requires that there must exist some value $x$ such that $\li{\sigma}{o_2}{\la{\rho}{o_2}}(x)=0$. If our output non-linearity does not have this property of having at least one zero in its range, then we cannot provide a zero initialization. For example, this means that we cannot use a sigmoid activation function on the output of our zero initialized network.

\subsection{\Rdeeper}

Given the derivation of zero initializations we just covered, the discussion in section \ref{sec:rdeeper} is a sufficient derivation, as given zero initializations \Rdeeper\ follows quickly. We still keep this section for completeness of appendix \ref{app:full_derivation}.

\section{Architectures and hyper-parameters}\label{app:hyperparam}

\begin{table*}[t]
    \caption{Definition of architectures used for the visualizations in figure \ref{fig:viz}.}
    \vskip 0.15in
    \begin{center}
    \begin{small}
    \begin{sc}
        \begin{tabular}{c|c|c|c}
            \toprule
                Layer Name      & Output Size   & SmallConv   & SmallConv(Widened) \\
            \midrule
                Conv1           & $32\times 32$ & $7\times 7$, 16 & $7\times 7$, 32 \\
            \hline
                Pool1           & $16\times 16$ & \multicolumn{2}{c}{$2\times 2$ Max Pool, stride 2} \\
            \hline
                Fc1           & $1\times 1$ & \multicolumn{2}{c}{150-D FC} \\
            \hline
                              & $1\times 1$ & \multicolumn{2}{c}{10-D FC} \\
            \bottomrule
        \end{tabular}
    \end{sc}
    \end{small}
    \end{center}
    \label{tab:viz_architectures}
    \vskip -0.1in
\end{table*}

\begin{table*}[t]
    \caption{Definition of architectures used for the tests on Cifar-10. We use the same notation as He \etal\ \yrcite{resnet}, where each block is defined in square brackets, each layer is specified by the kernel size and number of filters, and there is a residual connection over every block. We use the floor function, $\lfloor \cdot \rfloor$, for the number of channels, to allow for non-integer $r$. All $3\times 3$ convolutions use a padding of 1, and $7\times 7$ convolutions use a padding of 3. All convolutions use a stride of one, except from \begin{sc}Conv1\end{sc}, \begin{sc}Conv2\_1\end{sc} and \begin{sc}Conv3\_1\end{sc} which use a stride of two.}
    \vskip 0.15in
    \begin{center}
    \begin{small}
    \begin{sc}
        \begin{tabular}{c|c|c|c}
            \toprule
                Layer Name      & Output Size   & ResNetCifar-10($r$)   & ResNetCifar-18($r$) \\
            \midrule
                Conv1           & $16\times 16$ & \multicolumn{2}{c}{$7\times 7$,\ 64r, \text{stride}\ 2} \\
            \hline
                \multirow{2}{*}{Conv2\_x}
                                & \multirow{2}{*}{$8\times 8$}
                                                & \multicolumn{2}{c}{$3\times 3$\ \text{max pool},\ \text{stride}\ 2}  \\
            \cline{3-4}
                                &               & \gape{$\begin{bmatrix}
                    3\times 3,\ \lfloor 64r \rfloor \\
                    3\times 3,\ \lfloor 64r \rfloor
                \end{bmatrix} \times 2$}                                 & \gape{$\begin{bmatrix}
                    3\times 3,\ \lfloor 64r \rfloor \\
                    3\times 3,\ \lfloor 64r \rfloor
                \end{bmatrix} \times 4$} \\
            \hline
                Conv3\_x       & $4\times 4$      & \gape{$\begin{bmatrix}
                    3\times 3,\ \lfloor 128r \rfloor \\
                    3\times 3,\ \lfloor 128r \rfloor
                \end{bmatrix} \times 2$}                                & \gape{$\begin{bmatrix}
                    3\times 3,\ \lfloor 128r \rfloor \\
                    3\times 3,\ \lfloor 128r \rfloor
                \end{bmatrix} \times 4$} \\
            \hline
                                & $1\times 1$      & \multicolumn{2}{c}{\text{Avg Pool, 10-d FC, Softmax}}  \\
            \bottomrule
        \end{tabular}
    \end{sc}
    \end{small}
    \end{center}
    \label{tab:cifar_architectures}
    \vskip -0.1in
\end{table*}

To be as transparent as possible with how we ran our experiments, in tables \ref{tab:viz_architectures} and \ref{tab:cifar_architectures} we provide descriptions of the ResNet architectures we used, using the same notation as \citeapp{resnet} in the original ResNet paper. Moreover, for every test run we provide all training parameters in table \ref{tab:hyperparams}. Apart from the \textit{teacher networks}, which were heavily regularize to avoid the problem described in section \ref{sec:net_converge_tests}, hyper-paramters were chosen using a hyper-parameter search for each network.

For all of our tests, inputs to the neural networks were color normalized. All of the training and validation accuracies computed for Cifar-10 were using a single view and the entire $32\times 32$ image as input, with no cropping. We performed no other augmentation.

In all tests, a widen operation consists of multiplying the number of filters by a factor of $1.5$ , therefore, when a widen operation is applied to ResNetCifar($\frac{1}{8}$) it is transformed to a ResNetCifar($\frac{3}{16}$) network. We only ever applied a deepen operation to ResNetCifar10(r), transforming it into a corresponding ResNetCifar18(r).

All remaining training details are specified as hyper-parameters in table \ref{tab:hyperparams}.

\newpage
\begin{table*}[ht]
    \caption{Here we specify all hyper-parameters used in our tests. In the experiment name we use the following naming convention: [Test]-[NetworkName]. For the network convergence tests, we abbreviate the prefix to NCT, and for the faster training tests, we abbreviate the prefix to FTT. The prefixes that we use for the visualizations, overfitting (figure \ref{fig:overfitting_results}) and bad initializations (figure \ref{fig:bad_weight_init_results}) are abbreviated to VIZ, OF and BI.}
    \vskip 0.15in
    \begin{center}
    \begin{small}
    \begin{sc}
        \begin{tabular}{cccccccc}
            \toprule
                Experiment Name     & (Starting) Architecture      & Batch Size    & Optimizer     & Weight Decay     \\
            \midrule
                Viz-R2R & SmallConv & 128 & Adam & $10^{-3}$ \\
                Viz-Net2Net & SmallConv & 128 & Adam & $10^{-3}$ \\
                Viz-NetMorpg & SmallConv & 128 & Adam & $10^{-3}$ \\
                OF-Teacher & ResNetCifar18($\frac{1}{8}$) & 128 & Adam & 0 \\
                OF-Student & OF-Teacher & 128 & Adam & $3\times 10^{-3}$ \\
                OF-Rand & ResNetCifar18($\frac{3}{16}$) & 128 & Adam & $3\times 10^{-4}$ \\
                BI-He & ResNetCifar18($\frac{1}{8}$) &128 & Adam & $10^{-3}$ \\
                BI-Std & ResNetCifar18($\frac{1}{8}$) &128 & Adam & $10^{-3}$ \\
                NCT-WidenTeacher & ResNetCifar18($\frac{1}{8}$) & 128 & Adam & $5\times 10^{-3}$ \\
                NCT-R2WiderR & NCT-WidenTeacher & 128 & Adam & $10^{-2}$ \\
                NCT-NetMorph & NCT-WidenTeacher & 128 & Adam & $10^{-2}$ \\
                NCT-RandPadWiden & NCT-WidenTeacher & 128 & Adam & $10^{-2}$ \\
                NCT-WidenTeacher(NoResidual) & ResNetCifar18($\frac{1}{8}$) & 128 & Adam & $5\times 10^{-3}$ \\
                NCT-Net2WiderNet & NCT-WidenTeacher(NoResidual) & 128 & Adam & $10^{-2}$ \\
                NCT-DeepenTeacher & ResNetCifar18($\frac{1}{8}$) & 128 & Adam & $5\times 10^{-3}$ \\
                NCT-R2DeeperR & NCT-DeeperTeacher & 128 & Adam & $10^{-2}$ \\
                NCT-RandPadDeeper & NCT-DeeperTeacher & 128 & Adam & $10^{-2}$ \\
                NCT-DeeperTeacher(NoResidual) & ResNetCifar18($\frac{1}{8}$) & 128 & Adam & $5\times 10^{-3}$ \\
                NCT-Net2DeeperNet & NCT-DeeperTeacher(NoResidual) & 128 & Adam & $10^{-2}$ \\
                FFT-R2WiderR & ResNetCifar18($\frac{1}{8}$) & 128 & Adam & $10^{-2}$ \\
                FFT-NetMorph & ResNetCifar18($\frac{1}{8}$) & 128 & Adam & $10^{-2}$ \\
                FFT-RandPadWiden & ResNetCifar18($\frac{1}{8}$) & 128 & Adam & $10^{-2}$ \\
                FFT-Net2WiderNet & ResNetCifar18($\frac{1}{8}$) & 128 & Adam & $10^{-2}$ \\
                FFT-R2DeeperR & ResNetCifar18($\frac{1}{8}$) & 128 & Adam & $10^{-2}$ \\
                FFT-RandPadDeeper &ResNetCifar18($\frac{1}{8}$) & 128 & Adam & $10^{-2}$ \\
                FFT-Net2DeeperNet & ResNetCifar18($\frac{1}{8}$) & 128 & Adam & $10^{-2}$ \\
            \bottomrule
        \end{tabular}
        \begin{tabular}{cccccccc}
            \toprule
                Experiment Name     &  Transformations & Learning Rate     & LR Drops & Epochs Trained  \\
            \midrule
                Viz-R2R & [Widen@150 epochs] & $10^{-3}$ & N/A & 600 \\
                Viz-Net2Net & [Widen@150 epochs] & $10^{-3}$ & N/A & 600 \\
                Viz-NetMorph & [Widen@150 epochs] & $10^{-3}$ & N/A & 600 \\
                OF-Teacher & N/A & $10^{-3}$ & N/A & 250 \\
                OF-Student & [Widen@0 epochs] & $10^{-3}$ & [$\frac{1}{5}$@0 epochs] & 250 \\
                OF-Rand & N/A & $10^{-3}$ & & 250 \\
                BI-He & [Widen@20 epochs] & $10^{-3}$ & [$\frac{1}{5}$@20 epochs] & 120 \\
                BI-Std & [Widen@20 epochs] & $10^{-3}$ & [$\frac{1}{5}$@20 epochs] & 120 \\
                NCT-WidenTeacher & N/A & $3\times 10^{-3}$ & N/A & 250 \\
                NCT-R2WiderR & [Widen@0 epochs] & $3\times 10^{-3}$ & [$\frac{1}{5}$@0 epochs] & 250 \\
                NCT-NetMorph & [Widen@0 epochs] & $3\times 10^{-3}$ & [$\frac{1}{5}$@0 epochs] & 250 \\
                NCT-RandPadWiden & [Widen@0 epochs] & $3\times 10^{-3}$ & [$\frac{1}{5}$@0 epochs] & 250 \\
                NCT-WidenTeacher(NoResidual) & N/A & $3\times 10^{-3}$ & N/A & 250 \\
                NCT-Net2WiderNet & [Widen@0 epochs] & $3\times 10^{-3}$ & [$\frac{1}{5}$@0 epochs] & 250 \\
                NCT-DeeperTeacher & N/A & $3\times 10^{-3}$ & N/A & 250 \\
                NCT-R2DeeperR & [Deepen@0 epochs] & $3\times 10^{-3}$ & [$\frac{1}{5}$@0 epochs] & 250 \\
                NCT-RandPadDeeper & [Deepen@0 epochs] & $3\times 10^{-3}$ & [$\frac{1}{5}$@0 epochs] & 250 \\
                NCT-DeeperTeacher(NoResidual) & N/A & $3\times 10^{-3}$ & N/A & 250 \\
                NCT-Net2DeeperNet & [Deepen@0 epochs] & $3\times 10^{-3}$ & [$\frac{1}{5}$@0 epochs] & 250 \\
                FFT-R2WiderR & [Widen@25 epochs] & $3\times 10^{-3}$ & [$\frac{1}{5}$@25 epochs] & 250 \\
                FFT-NetMorph & [Widen@25 epochs] & $3\times 10^{-3}$ & [$\frac{1}{5}$@25 epochs] & 250 \\
                FFT-RandPadWiden & [Widen@25 epochs] & $3\times 10^{-3}$ & [$\frac{1}{5}$@25 epochs] & 250 \\
                FFT-Net2WiderNet & [Widen@25 epochs] & $3\times 10^{-3}$ & [$\frac{1}{5}$@25 epochs] & 250 \\
                FFT-R2DeeperR & [Deepen@25 epochs] & $3\times 10^{-3}$ & [$\frac{1}{5}$@25 epochs] & 250 \\
                FFT-RandPadDeeper & [Deepen@25 epochs] & $3\times 10^{-3}$ & [$\frac{1}{5}$@25 epochs] & 250 \\
                FFT-Net2DeeperNet & [Deepen@25 epochs] & $3\times 10^{-3}$ & [$\frac{1}{5}$@25 epochs] & 250 \\
            \bottomrule
        \end{tabular}
    \end{sc}
    \end{small}
    \end{center}
    \label{tab:hyperparams}
    \vskip -0.1in
\end{table*}

\FloatBarrier

\section{Weight initialization in \R} \label{apx:bad_weight_init}

\begin{figure}[h]
    \centering
    \includegraphics[width=0.7\columnwidth]{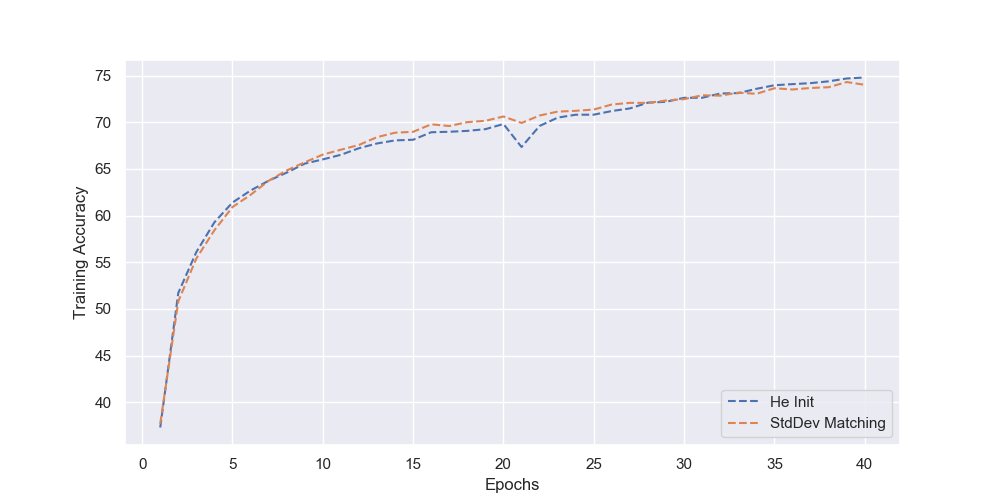} 
    \caption{Two networks widened at 20 epochs. Weights using He initialization are much larger than existing parameters, and causes some instability if used. We can observe similar effects with deepening.}
    \label{fig:bad_weight_init_results}
\end{figure}

In figure \ref{fig:bad_weight_init_results} we compare what happens if we use a more standard He initialization for new parameters in the network, compared to matching the standard deviations of local weights as described in section \ref{sec:faster_convergence}.

\section{Future work}



Although we have discussed \R, \N\ and Network Morphism in depth in this work, there are still some properties that we would have liked to consider. For example, we typically need to change the learning rate and weight decay, and be careful about the initialization scale of new parameters. We would have liked to consider the interplay between the learning rate, weight decay, weight initialization and optimizer, and provided (theoretical) guidance on how to manipulate these hyper-parameters best when performing an FPT.

Additionally, we think that the freedom in the initializations could allow more complex iniilization schemes to be considered. For example, a hybrid system between \Rwider\ and \Nwider\ could be considered, where new channels in \Rwider are initialized as copies of other channels. Alternatively, learning good initializations has been considered in the meta-learning problem \citeapp{finn2017model}, and so maybe `good' initializations could be learned in some problems.

Finally, Network Morphism also considers kernel size morphing and subnet morphing, and we have not consdered these transformations in this work. We note that our zero initialization from section \ref{sec:zero_init} could be used as an alternative subnet morphing scheme to that considered by \citeapp{netmorph}.

\newpage

\bibliographystyleapp{icml2019}
\bibliographyapp{main.bib}

\end{document}